\documentclass[sigconf]{acmart}
\usepackage{bm}
\usepackage{fontawesome} 
\usepackage{xcolor}      
\usepackage{makecell}

\AtBeginDocument{%
  }

\begin{document}


\title{Generalizable Engagement Estimation in Conversation via Domain Prompting and Parallel Attention}

\author{Yangchen Yu}
\authornote{Equal contribution.}
\affiliation{%
  \institution{School of Computer Science and Information Engineering, Hefei University of Technology}
  \city{Hefei}
  \country{China}}
\email{2019212292@mail.hfut.edu.cn}

\author{Yin Chen}
\authornotemark[1]
\affiliation{%
  \institution{School of Computer Science and Information Engineering, Hefei University of Technology}
  \city{Hefei}
  \country{China}}
\email{chenyin@mail.hfut.edu.cn}

\author{Jia Li}
\authornote{Corresponding author.}
\affiliation{%
  \institution{School of Computer Science and Information Engineering, Hefei University of Technology}
  \city{Hefei}
  \country{China}}
\email{jiali@hfut.edu.cn}

\author{Peng Jia}
\affiliation{%
  \institution{School of Computer Science and Information Engineering, Hefei University of Technology}
  \city{Hefei}
  \country{China}}
\email{2020214631@mail.hfut.edu.cn}

\author{Yu Zhang}
\affiliation{%
  \institution{School of Computer Science and Information Engineering, Hefei University of Technology}
  \city{Hefei}
  \country{China}}
\email{yuyueback@gmail.com}

\author{Li Dai}
\affiliation{%
  \institution{School of Computer Science and Information Engineering, Hefei University of Technology}
  \city{Hefei}
  \country{China}}
\email{321daili123@gmail.com}

\author{Zhenzhen Hu}
\affiliation{%
  \institution{School of Computer Science and Information Engineering, Hefei University of Technology}
  \city{Hefei}
  \country{China}}
\email{zzhu@hfut.edu.cn}

\author{Meng Wang}
\affiliation{%
  \institution{School of Computer Science and Information Engineering, Hefei University of Technology}
  \city{Hefei}
  \country{China}}
\email{eric.mengwang@gmail.com}

\author{Richang Hong}
\affiliation{%
  \institution{School of Computer Science and Information Engineering, Hefei University of Technology}
  \city{Hefei}
  \country{China}}
\email{hongrc.hfut@gmail.com}

\renewcommand{\shortauthors}{Yu, Chen, and Li et al.}

\begin{abstract}
Accurate engagement estimation is essential for adaptive human-computer interaction systems, yet robust deployment is hindered by poor generalizability across diverse domains (e.g., cultures and languages) and challenges in modeling complex interaction dynamics.
To tackle these issues, we propose \textbf{DAPA} (\textbf{D}omain-\textbf{A}daptive \textbf{P}arallel \textbf{A}ttention), a novel framework for generalizable conversational engagement modeling.  DAPA introduces a Domain Prompting mechanism by prepending learnable domain-specific vectors to the input, explicitly conditioning the model on the data’s origin to facilitate domain-aware adaptation while preserving generalizable engagement representations. To capture interactional synchrony, the framework also incorporates a Parallel Cross-Attention module that explicitly aligns reactive (forward BiLSTM) and anticipatory (backward BiLSTM) states between participants.
 Extensive experiments demonstrate that DAPA establishes a new state-of-the-art performance on several cross-cultural and cross-linguistic benchmarks, notably achieving an absolute improvement of 0.45 in Concordance Correlation Coefficient (CCC) over a strong baseline on the NoXi-J test set. The superiority of our method was also confirmed by winning the first place in the Multi-Domain Engagement Estimation Challenge at MultiMediate'25. The source code will be made available at https://github.com/MSA-LMC/DAPA.

\end{abstract}





\maketitle

\section{Introduction}

The quest for truly empathetic and collaborative AI systems hinges on their ability to perceive human engagement—the degree of involvement one expresses in a conversation \cite{sidner2002human, li2024dat, graffigna2015measuring}. Engagement is not a static attribute but a dynamic, interactive phenomenon, conveyed through a rich tapestry of multimodal signals such as gaze, nods, and vocal backchannels \cite{reece2023candor,muller2022multimediate}. Yet, the path towards creating robust and generalizable engagement models is impeded by two fundamental obstacles.

Chief among these is the profound \textit{domain gap} stemming from cultural and linguistic diversity \cite{9737677, jung2022great,chen2021temporal}. The communicative function of behaviors like head nods or pause durations can vary dramatically across cultures, causing models trained in one context to fail when deployed in another. This gap severely restricts the real-world utility of current systems.

Compounding this issue is a persistent oversight in modeling the \textit{interpersonal dynamics} of engagement \cite{koul2023spontaneous,wohltjen2024interpersonal}. While widely acknowledged as an emergent property of interaction \cite{dermouche2019engagement}, many models reduce engagement to an individual's state, neglecting the crucial influence of their conversational partners \cite{kumar2024towards, yu2023sliding}. Even recent attempts to incorporate partner information through simple cross-attention \cite{li2024dat} fall short, failing to capture the fine-grained, moment-to-moment synchrony that underpins genuine connection \cite{koul2023spontaneous}.

In response to these intertwined challenges, we introduce the \textbf{Domain-Adaptive Parallel Attention Network (DAPA)}, a novel framework designed for robust and interaction-aware engagement modeling. DAPA confronts the domain gap with a \textbf{Domain Prompting} mechanism, prepending learnable vectors to explicitly condition the model on the data's cultural origin. To unravel complex interaction dynamics, it employs a \textbf{Parallel Cross-Attention} module. This module uniquely aligns participants' \textit{Reactive} (forward BiLSTM) and \textit{Anticipatory} (backward BiLSTM) states, thereby capturing the delicate dance of moment-to-moment synchrony.

The efficacy of DAPA is demonstrated not only by achieving state-of-the-art performance on multiple challenging cross-cultural benchmarks but also by securing the first place in the Multi-Domain Engagement Estimation Challenge \cite{multimediate25_engagement} at MultiMediate'25. Our main contributions are as follows:

\begin{enumerate}
    \item \textbf{A novel domain prompting mechanism} that enables robust cross-domain generalization by prepending learnable domain-specific vectors, allowing the model to condition on dataset-specific engagement patterns during joint training.
    \item \textbf{A parallel cross-attention module} that explicitly models interaction dynamics by aligning the target participant's reactive and anticipatory BiLSTM states with those of their conversational partner, capturing fine-grained moment-to-moment synchrony.
    \item \textbf{Extensive validation of state-of-the-art performance} across multiple benchmarks, exemplified by an absolute CCC improvement of 0.45 on the NoXi-J test set, demonstrating the model's superior effectiveness and generalizability.
\end{enumerate}

\section{Related Works}
\textbf{Multimodal and Cross-Domain Engagement Estimation.} Accurately estimating participant engagement is fundamental to understanding human interaction. Prior works have explored this task from various modalities and perspectives \cite{yu2004detecting,monkaresi2016automated,chen2019faceengage,karimah2022automatic,stappen2022estimation}, yet few have considered the implications of cultural diversity \cite{liu2023cross,gao2023cross,xie2025multimodal}, which plays a pivotal role in psychological and behavioral analysis \cite{elfenbein2002universality}. Rudovic et al. \cite{rudovic2017measuring,rudovic2018culturenet} conducted preliminary investigations into how engagement varies across cultural contexts, highlighting the need for cross-domain generalization. While related domains such as sentiment analysis have introduced cross-domain adaptation techniques—e.g., DiSRAN \cite{9718029} employing adversarial disentanglement, and Wang et al. \cite{10606039} using feature decoupling via linear transformations—such strategies remain underexplored in engagement estimation. Existing methods \cite{yu2023sliding,li2024dat} typically require dataset-specific training and exhibit poor generalizability across domains. To address this limitation, we propose a novel Domain Prompting mechanism that enables dynamic domain adaptation within a unified framework. Our method demonstrates consistent performance gains and achieves state-of-the-art results across multiple culturally and linguistically diverse datasets.

\begin{figure*}[!ht]
  \centering
  \includegraphics[width=0.95 \linewidth]{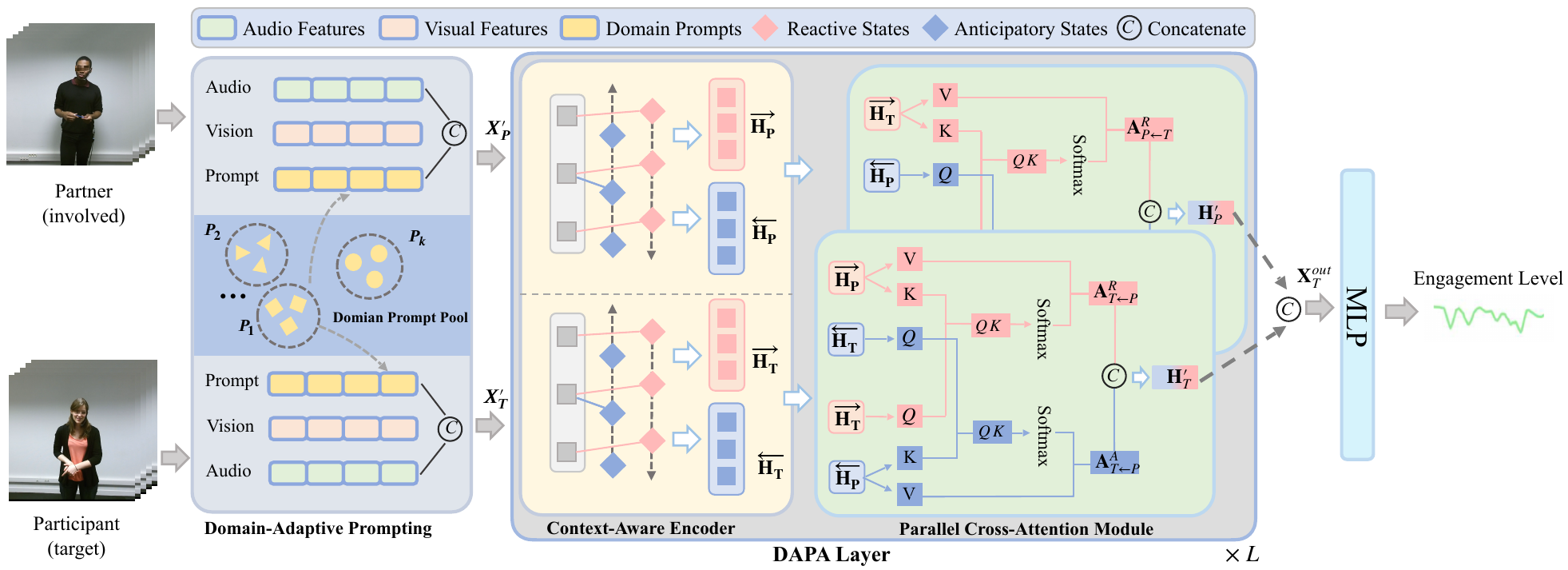}
  \caption{Overview of the DAPA framework. DAPA introduces two key innovations: (a) Domain Prompting, which prepends learnable domain-specific vectors to input features for cross-domain adaptation, and (b) a stack of L DAPA Layers. Each layer models interaction by employing a Parallel Cross-Attention mechanism that aligns the Reactive (forward BiLSTM) and Anticipatory (backward BiLSTM) states between the target and their partner. The figure illustrates the attention flow for the target participant; a symmetric operation is performed for the partner. Finally, the deeply interactive representations are fed into a prediction head to estimate the engagement score.}
  \label{fig:model_noxi}
\end{figure*}

\textbf{Modeling Interaction Dynamics.} Engagement is an emergent property of interpersonal dynamics, yet many estimation models \cite{ishii2008estimating,ishii2013gaze,ooko2011estimating,pellet2023multimodal} fail to capture this interactive essence. Early approaches adopted a participant-centric view, predicting engagement from an individual's features in isolation \cite{yu2023sliding}, thereby disregarding the reciprocal influence fundamental to conversation. While subsequent methods like DAT \cite{li2024dat} acknowledged this gap by incorporating partner information, they relied on shallow fusion techniques such as simple cross-attention \cite{tang2025asyrec,stamate2024ensembles}. This approach is conceptually limited, as it treats interaction as a static aggregation of signals rather than a dynamic process, failing to model the complex, non-linear dependencies and temporal synchrony that characterize human interaction. In contrast, our proposed DAPA employed a parallel attention architecture designed for deep relational modeling, explicitly capturing fine-grained, moment-to-moment synchrony and interdependence between participants.

\section{Methodology}

\subsection{Task Formulation}

Given a cross-cultural, multi-domain conversation corpus \( D \), we represent it as a set of samples \( \{x_1, x_2, \dots, x_D\} \). Each sample comprises an audio-visual sequence \( S = \{S_1, S_2, \dots, S_N\} \) spanning \( N \) frames. Each frame \( S_i = \{S_i^a, S_i^v\} \)  contains synchronized features from two modalities: audio ($a$) and visual ($v$). The objective of the engagement estimation task is to learn a mapping $f: S_i \mapsto y_i$, where $y_i \in [0, 1]$ denotes the continuous engagement score for frame $S_i$. Accordingly, the model predicts a sequence of engagement scores $\mathbf{Y} = \{y_1, y_2, \dots, y_N\}$ for the input sequence $S$. Notably, the corpus encompasses long-duration conversations across multiple languages, formats, and topical domains, introducing significant variation in both content and interaction style.

\subsection{Overall Architecture}
Our proposed Domain-Adaptive Parallel Attention framework (DAPA) is designed to accurately predict the engagement of a target participant ($P_T$) by leveraging multimodal information from both the target and their conversational partner ($P_P$). As depicted in Figure \ref{fig:model_noxi}, the DAPA pipeline begins by enriching input features with \textbf{Domain Prompts} to handle cross-domain variability. The core of the model consists of a stack of identical DAPA Layers, where each layer refines the participant's representation through a novel \textbf{Parallel Cross-Attention Module}. This module explicitly models the interaction by aligning the \textbf{Reactive} and \textbf{Anticipatory} states between the participants. Finally, a prediction head maps the deeply interactive features to a continuous engagement score. The following sections will detail each component of this architecture.

\subsection{Domain-Adaptive Prompting}
The initial stage of DAPA focuses on constructing a rich, domain-aware input representation for both the target and partner.

\subsubsection{\textbf{Multimodal Feature Extraction}}
For each participant, we extract a comprehensive set of features from their audio-visual data streams $S$. From the audio modality, we combine low-level acoustic descriptors (eGeMAPS\cite{eyben2010opensmile}) with high-level semantic features from a pretrained Whisper \cite{radford2023robust} model. For the visual modality, we fuse global scene context (Swin Transformer\cite{liu2021Swin}) with fine-grained behavioral cues from facial landmarks (OpenFace\cite{baltruvsaitis2016openface}) and body keypoints (OpenPose\cite{cao2017realtime}). These multimodal features are then projected and concatenated into a frame-level feature sequence, $\mathbf{X}\in \mathbb{R}^{N\times D_{in}}$, for both the target participant ($\mathbf{X_T}$) and the partner ($\mathbf{X_P}$).
\subsubsection{\textbf{Domain prompting}}
To bridge the significant domain gap present in cross-cultural and cross-lingual datasets, we introduce a \textit{Domain Prompting} mechanism. Instead of forcing the model to learn domain-invariant features, which may discard valuable domain-specific information, we explicitly condition the model on the data's origin.

Specifically, we construct a Domain Prompt Pool (DPP), denoted as $ P = \{P_1, P_2, \dots, P_k\} $, where each vector $p_d \in \mathbb{R}^{N\times D_p}$ corresponds to one of the $K$ domains (datasets) in our training corpus. 
These prompts are randomly initialized and optimized end-to-end with the rest of the network. For a given input sequence from domains (datasets) $d$, we prepend its corresponding prompt vector $p_d$ to the feature sequences of both the target participant and the partner:

\begin{equation}
    \mathbf{X'_T} = Concat(p_d, \mathbf{X_T}), \mathbf{X'_P} = Concat(p_d, \mathbf{X_P}). 
\end{equation}
This simple yet effective technique acts as an explicit instruction, allowing the network to activate different pathways or adjust its parameters to handle domain-specific engagement patterns, thereby enhancing its generalization capability.

\subsection{Reactive-Anticipatory Interaction Modeling}
To capture the intricate dynamics of a conversation, we introduce a novel interaction modeling approach. Central to this approach is the DAPA layer, which is stacked to form a hierarchical architecture. Each layer employs a Parallel Cross-Attention Module to explicitly model the synchrony between participants' reactive states and anticipatory contexts.

\subsubsection{\textbf{Context-Aware Encoder}}

Each DAPA layer first processes the input feature sequences of the target participant and the involved partner through independent BiLSTM encoders to capture temporal dependencies. A pivotal design choice in our architecture is the conceptual decomposition of the BiLSTM's output. Instead of using the concatenated hidden states monolithically, we separate them to distinguish between immediate, reactive behaviors and a holistic, anticipatory understanding of the conversation. Specifically, we define two distinct contextual representations. The first is the \textbf{Reactive States ($\overrightarrow{\mathbf{H}}$)}, composed of the hidden states from the forward pass, $\overrightarrow{\mathbf{H}} = \{\overrightarrow{h}_1, ..., \overrightarrow{h}_N\}$. Each state $\overrightarrow{h}_t$ models a participant's response to past and current events, thus representing their "in-the-moment" reactions. The second is the \textbf{Anticipatory Context ($\overleftarrow{\mathbf{H}}$)}, which consists of the hidden states from the backward pass, $\overleftarrow{\mathbf{H}} = \{\overleftarrow{h}_1, ..., \overleftarrow{h}_N\}$. Each state $\overleftarrow{h}_t$ is informed by the entire future of the conversation, thereby encoding a more global, "anticipatory" perspective. By applying these encoders to both the target and partner inputs, we obtain their respective reactive states ($\overrightarrow{\mathbf{H_T}}, \overrightarrow{\mathbf{H_P}}$) and anticipatory representations ( $\overleftarrow{\mathbf{H_T}}, \overleftarrow{\mathbf{H_P}}$), which are essential for the subsequent interaction modeling stage.

\subsubsection{\textbf{Parallel Cross-Attention Module}}
\label{sec:parallel_attention}

To model the reciprocal nature of interaction, we introduce the \textbf{Parallel Cross-Attention Module}. This module moves beyond simplistic, unidirectional information fusion by creating a bidirectional exchange that simultaneously refines both participants' representations. It operates through two parallel pathways, each modeling a different facet of interpersonal synchrony. The first pathway, \textbf{Anticipatory Context Alignment}, aims to model high-level contextual synchrony and shared understanding. It assesses how each participant's holistic, future-informed perspective aligns with the other's.   This is achieved through a bidirectional cross-attention mechanism. The target's anticipatory states $\overleftarrow{\bm{H}}_\text{T}$ query the partner's states $\overleftarrow{\bm{H}}_\text{P}$ (which serve as Key and Value), and simultaneously, the partner's states $\overleftarrow{\bm{H}}_\text{P}$ query the target's states $\overleftarrow{\bm{H}}_\text{T}$:

\begin{align}
    \bm{A}_{\text{T} \leftarrow \text{P}}^{A} &= \text{Attention}(\overleftarrow{\bm{H}}_\text{T}, \overleftarrow{\bm{H}}_\text{P}, \overleftarrow{\bm{H}}_\text{P}) \label{eq:att_T_anticip} \\
    \bm{A}_{\text{P} \leftarrow \text{T}}^{A} &= \text{Attention}(\overleftarrow{\bm{H}}_\text{P}, \overleftarrow{\bm{H}}_\text{T}, \overleftarrow{\bm{H}}_\text{T}) \label{eq:att_P_anticip}
\end{align}
where the \(\text{Attention}(Q, K, V)\) function is the standard scaled dot-product attention~\cite{vaswani2017attention}:
\begin{equation}
    \text{Attention}(Q, K, V) = \text{softmax}\left(\frac{QK^T}{\sqrt{d_k}}\right)V
    \label{eq:attention_def}
\end{equation}

The second pathway, \textbf{Reactive State Alignment}, focuses on capturing fine-grained, immediate behavioral synchrony, such as responsive gestures or mimicry.  It aligns each participant's ``in-the-moment'' reactions with the other's reactive states using the same bidirectional attention structure
\begin{align}
    \bm{A}_{\text{T} \leftarrow \text{P}}^{R} &= \text{Attention}(\overrightarrow{\bm{H}}_\text{T}, \overrightarrow{\bm{H}}_\text{P}, \overrightarrow{\bm{H}}_\text{P}) \label{eq:att_T_react} \\
    \bm{A}_{\text{P} \leftarrow \text{T}}^{R} &= \text{Attention}(\overrightarrow{\bm{H}}_\text{P}, \overrightarrow{\bm{H}}_\text{T}, \overrightarrow{\bm{H}}_\text{T}) \label{eq:att_P_react}
\end{align}

The output of this module consists of four interaction-aware representations that capture the mutual influence from both global and local perspectives. To prepare the inputs for the subsequent DAPA layer, we update each participant's representation by concatenating their respective reactive and anticipatory alignment vectors. For the target participant (T), this new representation, $\bm{H}'_\text{T}$, is formed as:

\begin{equation}
    \bm{H}'_\text{T} = \operatorname{Concat}\bigl( \bm{A}_{\text{T} \leftarrow \text{P}}^{R}, \bm{A}_{\text{T} \leftarrow \text{P}}^{A} \bigr)
    \label{eq:update_T}
\end{equation}
Similarly, the partner's representation (P) is updated to $\bm{H}'_\text{P}$:

\begin{equation}
    \bm{H}'_\text{P} = \operatorname{Concat}\bigl( \bm{A}_{\text{P} \leftarrow \text{T}}^{R}, \bm{A}_{\text{P} \leftarrow \text{T}}^{A} \bigr)
    \label{eq:update_P}
\end{equation}

These fused, interaction-aware representations, $\bm{H}'_\text{T}$ and $\bm{H}'_\text{P}$, are subsequently fed into the next DAPA layer, enabling hierarchical refinement of relational dynamics.

\subsection{\textbf{Engagement Prediction}}
\label{sec:prediction}


After processing through all $L$ DAPA layers, we obtain the final interaction-aware representations $\bm{H}'_\text{T}$ and $\bm{H}'_\text{P}$ for the target and partner, respectively (Eq.~\ref{eq:update_T},~\ref{eq:update_P}).
To predict the target's engagement, we first concatenate these representations to form a dyadic state vector,
\begin{equation}
    \bm{X}_\text{T}^{\text{out}} = \operatorname{Concat}\bigl( \bm{H}'_\text{T}, \bm{H}'_\text{P} \bigr),
    \label{eq:final_concat}
\end{equation}
which captures the full interactive context. A multi-layer perceptron (MLP)~\cite{rumelhart1986learning} with a Sigmoid activation then maps this joint representation to the final predicted engagement sequence:
\begin{equation}
    \hat{\bm{Y}}_T = \text{Sigmoid}\left(\text{MLP}(\bm{X}_T^{\mathrm{out}})\right).
    \label{eq:prediction_head}
\end{equation}

The entire DAPA model is trained end-to-end by minimizing the following loss function:
\begin{equation}
\mathcal{L} = 1 - \mathrm{CCC}(\hat{\mathbf{Y}}_T,\mathbf{Y}_T),
\label{eq:loss}
\end{equation}
where $\mathbf{Y}_T$ is the ground-truth engagement sequence. This objective directly aligns the training process with our primary evaluation metric, the Concordance Correlation Coefficient (CCC).

\section{Experiments}


\subsection{Datasets}
The challenge comprises four distinct datasets, and the final competition metric is the average CCC score across these four sets.

\textbf{NoXi Base \cite{cafaro17_icmi} dataset} contains 48 training sessions and 16 one-on-one video interaction test sessions between experts and novices conducted in English, French and German. Each session includes synchronized audio and visual streams recorded at 25 fps (with some features sampled at 40fps), totaling 84 dialogues involving 87 participants (26 female, 61 male) and roughly 25 hours of audio. Frame‑wise engagement scores in [0,1] provided for each frame are rated by 2–7 annotators (on average 3.6). For the 2025 Multi‑Domain Engagement Estimation Challenge, both pre‑extracted features and the raw video files were made available.

\textbf{NoXi Additional \cite{cafaro17_icmi, muller2024multimediate} Language dataset} (NoXi-Add) comprises only 12 test sessions. To assess cross‑cultural generalization beyond the core NoXi languages, this set introduces four new languages: Arabic, Italian, Indonesian, and Spanish, while maintaining the same feature formats as NoXi Base.

\textbf{NoXi Japan \cite{funk2024multilingual} dataset} (NoXi‑J) comprises 10 test sessions of expert–novice screen‑mediated interactions in Japanese and Mandarin Chinese, recorded in Japan and China under the original NoXi protocol. This expands coverage to East Asian languages beyond the eight European languages in the core NoXi.

\textbf{MPIIGroupInteraction \cite{mueller21_mm, muller2018detecting} dataset} (MPIIGI) comprises 6 test sessions recorded in an office setting using an 8 camera (4DV) array and four ambient microphones. Each 20‑minute session captures four participants discussing a self‑chosen controversial topic in German. Unlike the mostly standing interactions in NoXi, participants in MPIIGI are mostly seated; feature sampling rates match those of NoXi.

    

\subsection{Experimental Setup}
We used the official pre-extracted features, totaling 1709 dimensions: 88-D eGeMAPSv2 audio, 768-D Swin Transformer vision, 714-D OpenFace facial and 139-D OpenPose pose features. In addition, we extracted 1280-D Whisper audio embeddings, which yielded a total of 2989 dimensions.

Across all experiments, we fixed the random seed to 40 and used the Adam optimizer with an initial learning rate of $5 \times 10^{-5}$. We linearly warmed up for the first 400 steps to $5 \times 10^{-5}$, then applied cosine annealing with $T_{\max}=10$. Training and validation batch sizes were set to 32 and 256, respectively, over 40 epochs. Following the context‑segmentation scheme in DAT, the total sequence length was 96 frames, comprising a 32‑frame core segment plus two 32‑frame auxiliary segments, with a sliding stride $s=32$.

In the model, each input feature vector is first mapped to a 512‑D representation via two fully connected layers, then fed into a three‑layer  BiLSTM~\cite{schuster1997bidirectional, graves2012long} encoder with hidden sizes of 512. To mitigate overfitting and improve generalization, we applied a dropout rate of 0.1 across all layers and employed an exponential moving average (EMA)~\cite{polyak1992acceleration} over the entire 40‑epoch training. All training and evaluation used the CCC as the criterion.

\begin{figure}[!t]
  \captionsetup{justification=raggedright, singlelinecheck=false} 
  \includegraphics[width=0.98 \linewidth]{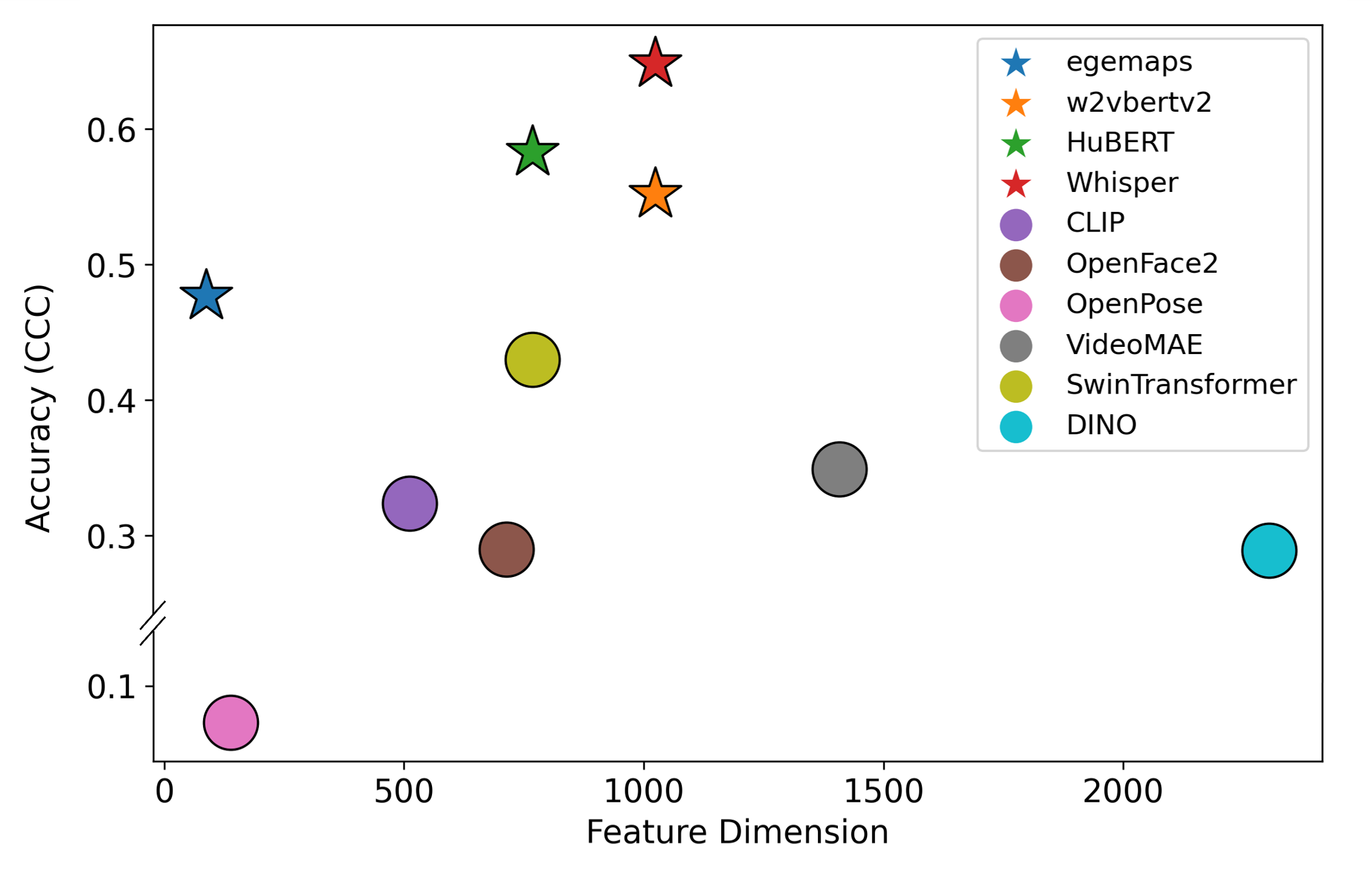}
\caption{Assessing the influence of key features on engagement estimation performance on the NoXi-J dataset. }
\label{fig:feature_dim_vs_accuracy}
\end{figure}

\subsection{Ablation Studies}
To verify the effectiveness of our proposed model, we conducted ablation experiments on the NoXi-J and NoXi-Base validation sets, focusing on two key components: Domain-Adaptive Prompt (DAP) and Parallel Cross-Attention (PCA). We also evaluated the contribution of each input feature. The baseline model without DAP and PCA is described in the first row of Table~\ref{tab:dap_pca_results}.

\begin{table}[h]
    \centering
    \caption{Ablation study results for the Domain‑Adaptive Prompt (DAP) and Parallel Cross‑Attention (PCA) configurations.}
    \begin{tabular}{@{}cccc@{}}
        \toprule
        \textbf{DAP} & \textbf{PCA} & \textbf{NoXi-J} & \textbf{NoXi-Base} \\
        \midrule
         &  & 0.669 & 0.822 \\
        \checkmark &  & 0.705 & 0.837 \\
        & \checkmark & 0.684 & 0.836 \\
        \checkmark & \checkmark & 0.722 & 0.855 \\
        \bottomrule
    \end{tabular}
    \label{tab:dap_pca_results}
\end{table}

\subsubsection{Ablation on Feature Selection} To evaluate the performance of different feature combinations, we thoroughly investigated the contribution of each component using our proposed DAPA framework. The results, presented in Figure \ref{fig:feature_dim_vs_accuracy}, guided our final feature selection. For audio features, the experiments revealed that Whisper features significantly outperformed Hubert~\cite{hsu2021hubert} and W2v-BERTv2\cite{chung2021w2v}. To capture emotional cues from audio, we also utilized eGeMAPS features. For video features, SwinTransformer proved most effective. Additionally, to incorporate explicit facial and pose information, we used facial landmarks and body keypoints obtained from OpenFace and OpenPose, respectively. Ultimately, this systematic evaluation led us to select a combination of five features that achieved optimal performance within the DAPA model.

\subsubsection{Ablation on Domain‑Adaptive Prompting} To demonstrate the effect of Domain-Adaptive Prompting (DAP), we compare model performance with and without this mechanism. As shown in Table~\ref{tab:dap_pca_results}, DAP consistently improves results across both datasets. On the cross-domain NoXi-J test set, it raises the CCC from 0.669 to 0.705 (a notable gain of 0.036), highlighting the effectiveness of domain conditioning in mitigating domain shifts. Even on the culturally mixed NoXi-Base, DAP yields a smaller yet meaningful improvement of 0.015, demonstrating its capacity to retain domain-specific cues. These results affirm DAP’s critical role in enhancing generalization across diverse, real-world conditions.

\subsubsection{Ablation on Parallel Cross-Attention Module}
Our Parallel Cross-Attention (PCA) module is designed to capture the rich, universal dynamics of interpersonal interaction. To verify its contribution, we compare a baseline BiLSTM against a model enhanced with PCA. The results in Table \ref{tab:dap_pca_results} show that PCA provides a solid performance gain, increasing the CCC by 0.015 on NoXi-J and 0.014 on NoXi-Base. This demonstrates that modeling the synchrony between reactive and anticipatory states is a fundamentally robust approach to understanding engagement.

Furthermore, the synergy between PCA and DAP is especially salient. Adding PCA to a DAP-equipped model (row 4 vs. row 2) yields a 0.017 CCC gain on NoXi-J, while introducing DAP atop PCA (row 4 vs. row 3) results in a larger boost of 0.038. This asymmetry suggests that PCA provides an interaction-aware feature basis that amplifies the effectiveness of domain-specific prompting. Their complementary roles underscore the necessity of combining both modules to achieve DAPA’s state-of-the-art performance.




\begin{figure*}[!ht]
  \centering
  \includegraphics[width=0.96 \linewidth]{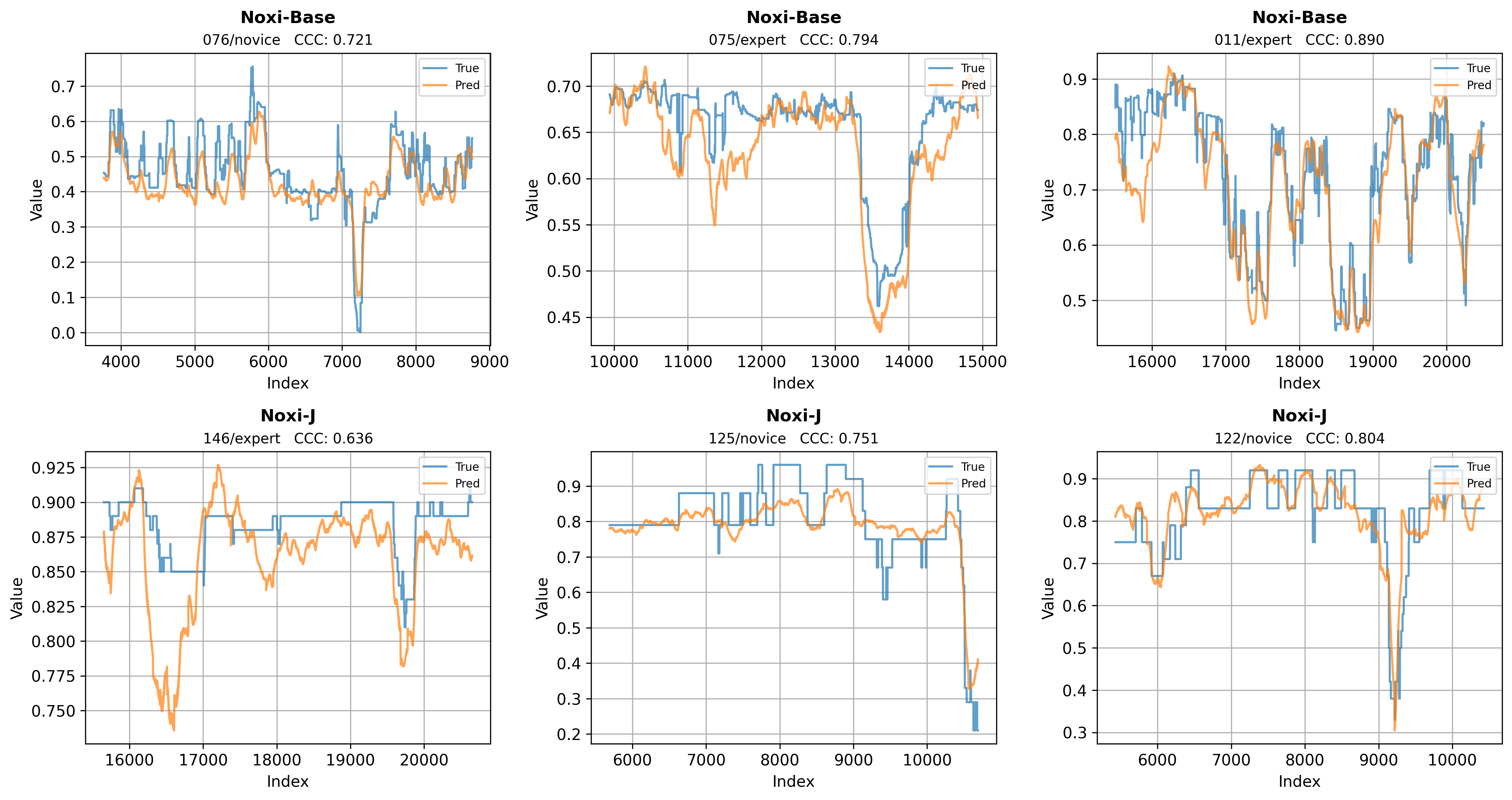}
  \caption{Real-time fitting results of the DAPA model on the NoXi-Base and NoXi-J datasets, with a fixed interval length of 5000 samples. True denotes the ground truth.}
  \label{fig:pred_vs_gt}
\end{figure*}

\definecolor{customgold}{RGB}{212, 175, 55}
\definecolor{customsilver}{RGB}{192, 192, 192}
\definecolor{custombronze}{RGB}{205, 127, 50}
\begin{table}[!t]
  \caption{Final CCC scores on various test datasets. Global represents the global average CCC across all datasets.}
  \label{tab:final_test}
  \resizebox{\linewidth}{!}{%
  \begin{tabular}{cccccc}
    \toprule
    \makecell{\textbf{Team}} & \makecell{\textbf{NoXi Base}} & \makecell{\textbf{NoXi-Add}} & \makecell{\textbf{MPIIGI}} & \makecell{\textbf{NoXi-J}} & \makecell{\textbf{Global}}\\
    \midrule
    \textbf{HFUT-LMC (Ours)}\textsuperscript{\textcolor{customgold}{\faTrophy}} & \textbf{0.79} & \textbf{0.75} & \textbf{0.67} & \textbf{0.58} & \textbf{0.70} \\
    USTC-IAT United\textsuperscript{\textcolor{customsilver}{\faTrophy}}  & 0.79 & 0.73 & 0.66  & 0.53 & 0.68 \\
    chfighting\textsuperscript{\textcolor{custombronze}{\faTrophy}}      & 0.77 & 0.70 & 0.65  & 0.50 & 0.65 \\
    nnaedu          & 0.75 & 0.68 & 0.57  & 0.58 & 0.64 \\
    lasii           & 0.79 & 0.73 & 0.54  & 0.51 & 0.64 \\
    yueangh         & 0.75 & 0.67 & 0.59  & 0.42 & 0.61 \\
    yueyue          & 0.65 & 0.67 & 0.61  & 0.34 & 0.57 \\
    MM25 Baseline \cite{cafaro17_icmi}   & 0.57 & 0.47 & 0.44 & 0.13 & 0.40 \\
    Behavioural-AI Lab  & 0.53 & 0.41 & 0.09  & 0.26 & 0.32 \\
    \bottomrule
  \end{tabular}}
\end{table}

\subsection{Final Results}
Finally, we applied our optimally configured framework to the test sets of all four datasets and compared them against the official baseline. As shown in Table \ref{tab:final_test} , our model achieved CCC scores of 0.79 on NoXi Base, 0.75 on NoXi‑Add, 0.58 on NoXi‑J, and 0.67 on MPIIGI, with a new state‑of‑the‑art. This not only demonstrates robustness but also highlights strong cross‑cultural and cross‑domain generalization. Overall, our approach outperforms the official baseline \cite{cafaro17_icmi} by 0.30 CCC on the global average metric, further validating its effectiveness.

\subsection{Visualization}


Figure~\ref{fig:pred_vs_gt} visualizes our DAPA model's predictions against the ground truth on the NoXi-Base (top row) and NoXi-J (bottom row) validation sets. 
A key observation is the stark difference in the ground truth characteristics between the two datasets. 
The NoXi-Base labels exhibit highly dynamic and continuous fluctuations, typical of fine-grained engagement annotation. 
In stark contrast, the NoXi-J labels follow a distinct step-like (quantized) pattern, characterized by abrupt transitions between sustained levels. 
Despite these fundamentally different annotation styles, our model's predictions (in orange) consistently and accurately track the ground truth (in blue) in both scenarios. 
This strongly demonstrates the model's robustness and its ability to generalize not just across subjects, but also across varying data distributions and annotation methodologies.

\section{Conclusion}
 
This paper introduces DAPA, a novel framework that significantly advances the state of the art in cross-cultural conversational engagement estimation. By synergizing a Domain Prompting mechanism for robust cross-domain adaptation with a Parallel Cross-Attention module that operates on learned reactive and anticipatory states, DAPA effectively captures the complex, interactive nature of engagement. The superiority of our model is demonstrated by establishing new state-of-the-art results on challenging benchmarks and securing first place in the MultiMediate'25 Engagement Estimation Challenge. This work thus provides a powerful and generalizable approach for multimodal interaction analysis.

Looking forward, we acknowledge certain limitations that offer clear paths for future inquiry. Our current Domain Prompting mechanism, while effective, requires retraining to adapt to new, unseen domains. Furthermore, our model's reliance on audio-visual features presents a major opportunity: the integration of linguistic content. Our exploratory experiments revealed a fundamental challenge in the temporal alignment between sparse textual utterances and dense, frame-level engagement annotations. We posit that developing sophisticated models to bridge this modality gap, such as dynamic attention architectures or asynchronous multimodal transformers, is a crucial next step. Successfully tackling these challenges will be key to unlocking a more holistic understanding of engagement, thereby paving the way for more contextually-aware and culturally-adaptive human-computer interaction systems.

\begin{acks}


\end{acks}

\newpage

\bibliographystyle{ACM-Reference-Format}
\balance
\bibliography{sample-base}

\appendix

\end{document}